\documentclass[letterpaper, 10 pt, conference]{ieeeconf}  
\usepackage{graphicx}
\usepackage{mdwlist}
\usepackage{fmtcount}
\usepackage{algorithmic}
\usepackage{algorithm} 
\usepackage{array}                                                         
\usepackage[tight,footnotesize]{subfigure}                                                          

\IEEEoverridecommandlockouts                              
\title{\LARGE \bf
Classification of Large-Scale Fundus Image Data Sets: A Cloud-Computing Framework
}

\author{Sohini Roychowdhury$^{1}$
\thanks{$^{1}$Department of Electrical Engineering, University of Washington, Bothell, WA 98011}%
}

\begin{document}

\maketitle
\thispagestyle{empty}
\pagestyle{empty}

\begin{abstract}
Large medical image data sets with high dimensionality require substantial amount of computation time for data creation and data processing. This paper presents a novel generalized method that finds optimal image-based feature sets that reduce computational time complexity while maximizing overall classification accuracy for detection of diabetic retinopathy (DR). First, region-based and pixel-based features are extracted from fundus images for classification of DR lesions and vessel-like structures. Next, feature ranking strategies are used to distinguish the optimal classification feature sets. DR lesion and vessel classification accuracies are computed using the boosted decision tree and decision forest classifiers in the Microsoft Azure Machine Learning Studio platform, respectively. For images from the DIARETDB1 data set, 40 of its highest-ranked features are used to classify four DR lesion types with an average classification accuracy of 90.1\% in 792 seconds. Also, for classification of red lesion regions and hemorrhages from microaneurysms, accuracies of 85\% and 72\% are observed, respectively. For images from STARE data set, 40 high-ranked features can classify minor blood vessels with an accuracy of 83.5\% in 326 seconds. Such cloud-based fundus image analysis systems can significantly enhance the borderline classification performances in automated screening systems.
\end{abstract}

{\bf Index Terms:} data mining; feature reduction; Microsoft Azure;

\section{Introduction}
Fundus images capture snapshots of the anterior portion of the eye, to detect retinal pathologies such as diabetic retinopathy (DR), glaucoma, macular edema, to name a few. Several automated diagnostic systems have been developed over the past decade that utilize fundus images for primary-care physicians to generate a quick ``second opinion'' and enable decision-making regarding referrals and follow-up treatment \cite{fraz} \cite{DREAM}. Most such automated diagnostic systems using fundus images are primarily based on machine learning and decision making principles. With increasing dimensions and sizes of medical data, automated decision making processes may experience scalability issues due to the speed, volume, variety and complexity involved with ``large-scale'' medical image data. In this paper, we present a scalable cloud-computing framework using Microsoft Azure Machine Learning Studio (MAMLS) platform to analyze and classify high-dimensional fundus image-based medical data sets and ensure high classification accuracy.

Large data sets with high dimensionality require substantial amount of computation time for data creation and data processing \cite{cloudref}. In such instances, data mining strategies such as feature reduction are found to be effective in enhancing manageability by significantly reducing the dimensionality and computational time complexity \cite{DREAM} cite{Major}. In this work a novel cloud-computing framework is presented that is capable of generalizing the steps for fundus image-based classification tasks to ensure maximum accuracy and low computational time complexity for automated DR screening systems. Most existing automated screening systems for non-proliferative DR (NPDR) ensure pathology detection at the cost of high false positives \cite{DREAM}. Proliferative DR (PDR) detection systems on the other hand, focus on retinal blood vessel extraction followed by classification for detection of new-vessel like abnormalities in the retina \cite{Lee}. All such automated DR detection systems primarily focus on classification accuracies per image, rather than the classification accuracy per lesion (or per pathological manifestation). The proposed system is trained to focus on pathology level classification to find generalizable features that discriminate borderline pathological manifestations from their normal counterparts. Such a generalized large-scale cloud-computing based analysis is capable of performing exhaustive feature set analysis and optimal classifier identification, thereby improving the state-of-the-art pathology classification metrics, thus leading to improved prognosis.

This paper makes two key contributions. First, it introduces a novel cloud-computing framework that processes large data sets to evaluate optimal classification features from fundus image data sets. This MAMLS generalized flow analyzes over 229,386 samples from fundus images with 98 features per sample by performing feature ranking, reduction and classification significantly in under 15 minutes of cloud-computing time. Second, several feature ranking strategies are comparatively analyzed and the minimal-redundancy-maximal-relevance (mRMR) \cite{mrmr} feature ranking strategy is found to be the best detector of optimal feature sets for fundus image classification tasks. The optimal feature sets are more discriminating than full feature sets. These optimal feature sets increase the overall classification accuracy from 0.2-1.2\% with 11-23\% reduction in computational time complexity when compared to the full feature set in the MAMLS platform.

\section{Data and Method}
This work analyzes the image-based features that uniquely identify retinal pathologies such as NPDR and blood vessel abnormalities due to PDR. While large numbers of image-based features can be useful in generalizing automated pathology classification methodologies, the identification of the optimal feature sets that maximize classification accuracies is key for accurate detection of the borderline pathological images. In this work, region-based and pixel-based features are analyzed for their impact on binary and multi-class classification for two separate automated pathology detection tasks based on the fundus image data sets described below. 
\subsection{Fundus Image Data}
\begin{figure}
\begin{center}
\subfigure[]{\includegraphics[width = 1.5in,height=0.9in]{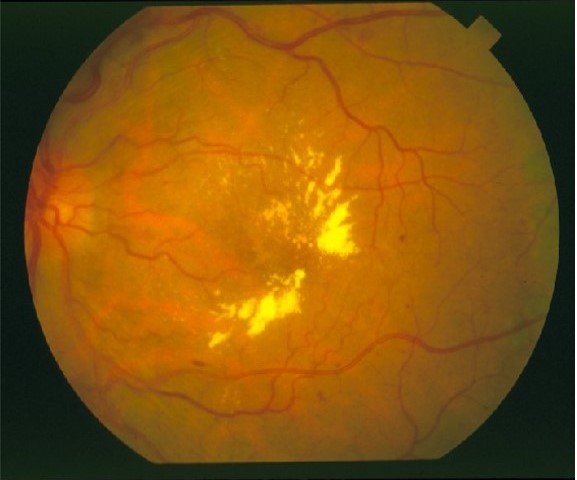}}
\subfigure[]{\includegraphics[width = 1.5in, height=0.9in]{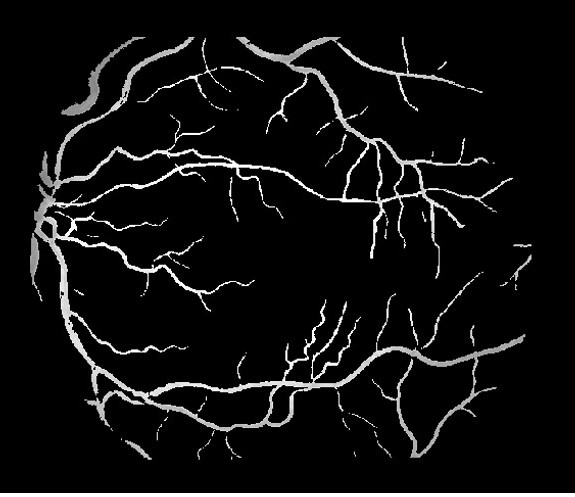}}
\subfigure[]{\includegraphics[width = 1.5in, height=0.9in]{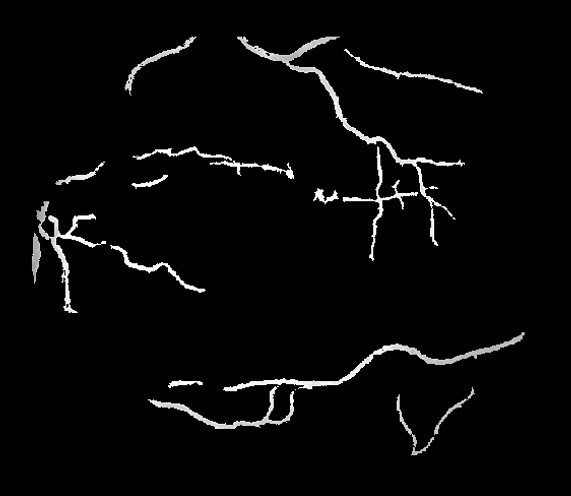}}
\subfigure[]{\includegraphics[width = 1.5in, height=0.9in]{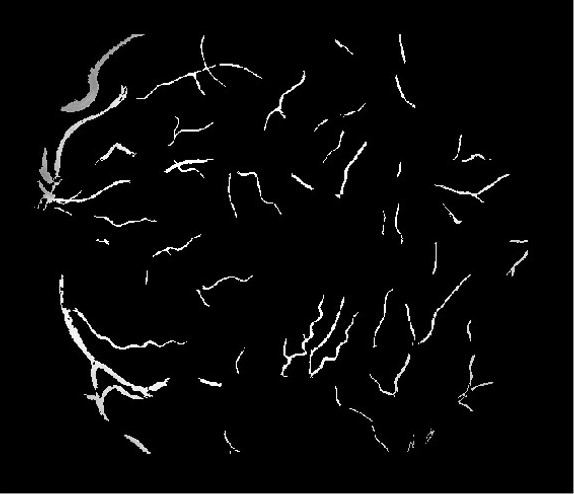}}
\caption{Blood vessel segmentation using fundus images (a) Fundus Image. (b) Manually marked blood vessels. (c) Major vessels detected using \cite{Major}. (d) Remaining minor vessel regions for binary classification.}     
\label{fundus}
\end{center}
\end{figure}

\begin{itemize}
\item DIARETDB1 \cite{DB1}: data set consists of 89 fundus images with $50^{o}$ FOV, that are manually annotated for bright lesions (hard exudates and cotton wool spots) and red lesions (haemorrhages and microaneurysms) corresponding to varying severities of NPDR. A sample image and the lesions are shown in Fig. \ref{DB1}. Automated image filtering and segmentation can be used to detect bright regions and red regions separately \cite{DREAM}, where each region corresponds to a sample for classification. An optimal set of region-based features corresponding to the bright and red regions can then be used to maximize the overall classification accuracy for such a multi-class classification task for NPDR detection with 6 classes (corresponding to false positive bright regions, hard exudates, cotton wool spots, false positive red regions, haemorrhages and microaneurysms, respectively).
\begin{figure}[ht]
\begin{center}
\includegraphics[width = 3.0in, height=1.75in]{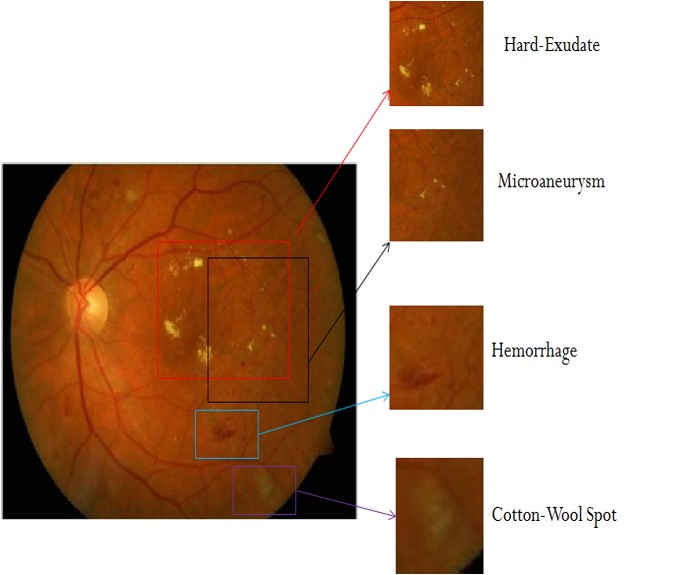}
\caption{A sample fundus image from DIARETDB1 data set with bright and red lesions corresponding to NPDR.}\label{DB1}
\end{center}
\end{figure}
\item STARE \cite{STARE}: data set contains 20 fundus images with 35$^{o}$ FOV that are manually annotated for blood vessels by two independent human observers. Here, 10 images represent patients with retinal abnormalities while the remaining 10 represent normal retina. A sample image and its vessel annotations are shown in Fig. \ref{fundus}(a),(b), respectively. Vessels marked by the second manual observer are considered ground-truth. PDR is known to cause fine vessel-like growth to appear in fundus images. Although the major blood vessel regions are easily detectable by high-pass and morphological filtering as shown in \cite{Major}, detection of finer vessel-like regions is challenging. An optimal set of region-based and pixel based features can then be used to classify the fine vessel regions from non-vessels (binary classification) to aid PDR detection. Here, each minor vessel region corresponds to a sample for classification.
\end{itemize}
 
\begin{figure}
\begin{center}
\subfigure[]{\includegraphics[width = 1.4in,height=1.0in]{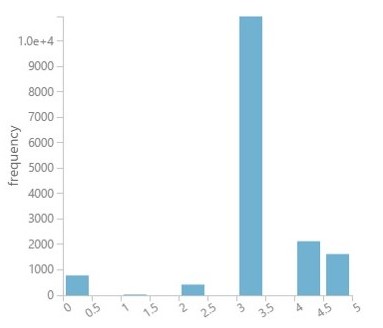}}
\subfigure[]{\includegraphics[width = 1.4in, height=1.0in]{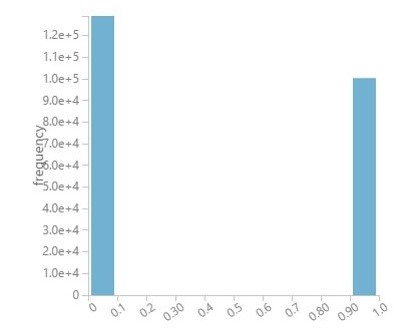}}
\caption{Sample distribution for the 2 data sets. (a) Distribution of the 6 class types from DIARETDB1 data set. False positive red lesions regions (class label 3) have the largest number of samples while cotton wool spot regions (class label 2) have the smallest number of samples. (b) Distribution of the 2 class types from the STARE vessel data set. Number of non-vessel regions (class label 0) is greater than the number of actual vessel regions (class label 1).}     
\label{hist}
\end{center}
\end{figure}

The histogram of sample distributions from our two data sets is shown in Fig. \ref{hist}. Classification of both data sets poses challenges due to the unbalanced sample distributions. Once the various sample regions are extracted from the fundus images, the next steps to extract features and classification are described in the sections below. 
\subsection{Feature Extraction}
The features that are extracted for classifying the region-based samples extracted from the data sets can be categorized into 7 categories shown in Table \ref{notation}. As a pre-processing step, the green plane of each fundus image is resized to [500x500] pixels and the pixel intensities are rescaled in the range [0,1], resulting in image $I$. From the RGB to HSI converted image planes \cite{DREAM}, the other similarly resized and rescaled image planes include the red plane ($I^r$), hue plane ($I^h$), saturation plane ($I^s$) and intensity plane ($I^i$). The Gaussian derivative images corresponding to 6 coefficients from 0th to second order Gaussian filtering of image $I$ in the horizontal ($x$ direction) and vertical ($y$ direction) with $\sigma^2=8$ are denoted as $[I^{G},I^{G}_x,I^{G}_y,I^{G}_{x,y},I^{G}_{xx},I^{G}_{yy}]$, respectively \cite{DREAM}. First and second order gradient images in $(x,y)$ directions for various image planes are denoted by the subscript $_{(x,y)}, _{(xx,yy)}$, respectively.
\begin{table}[ht]
\begin{center}
\caption{Definition of Features.}
\begin{tabular}{|c|l|l|}
\hline 
\#&Category&Features\\ \hline \hline
14&Structural&Area, bounding box lengths, convex area,\\
&&filled area, Euler number, extent, orientations,\\
&& major and minor axes lengths, orientation,\\
&&eccentricity, perimeter, solidity.\\\hline
12&Gaussian&Mean and variance in Gaussian coefficient\\
&Coefficients&images $[I^{G},I^{G}_x,I^{G}_y,I^{G}_{xy},I^{G}_{xx},I^{G}_{yy}]$.\\ \hline
16&Regional&Regional Mean, minimum, maximum and\\
&Intensity&std. dev. for images [$I,I^r,I^h,I^i$]\\\hline
24&Gradient&Maximum, minimum and mean pixel intensities\\
&Intensity&in gradient images $[I_{(x,y)},I_{(xx,yy)},I^{r}_{(x,y)},$\\
&&$I^{r}_{(xx,yy)},I^{h}_{(x,y)},I^{h}_{(xx,yy)},I^{s}_{(x,y)},I^{s}_{(xx,yy)}]$\\ \hline
24&Gradient in &Maximum, minimum and mean pixel intensities\\
&Image&in $[I.I_{(x,y)},I.I_{(xx,yy)},I^{r}.I^{r}_{(x,y)},I^{r}.I^{r}_{(xx,yy)},$\\
&Intensity&$I^{h}.I^{h}_{(x,y)},I^{h}.I^{h}_{(xx,yy)},I^{s}.I^{s}_{(x,y)},I^{s}.I^{s}_{(xx,yy)}]$\\ \hline
4&Pixel-window&Pixel intensity: Max. in [3x3], mean in [5x5], \\
&based \cite{Major}&std. dev. in [5x5], neighbors in [5x5] window.\\ \hline
4&Pixel intensity&From images $[I^{G}_x,I^{G}_y,I^{G}_{xx},I^{G}_{yy}]$.\\ \hline
\end{tabular}
	\label{notation}
	\end{center}
\end{table}
For the DIARETDB1 data set, $n=15,945$ samples with $L=66$ region-based features per sample are extracted using the 14, 12, 16 and 24 features corresponding to Structural, Gaussian Coefficient, Regional intensity and Gradient Intensity in Table \ref{notation}, respectively. For the STARE data set, $n=229,386$ samples with $L=98$ region-based and pixel-based features per sample are extracted using all the features defined in Table \ref{notation}. The next step is identification of the most discriminating features for classification tasks.

\subsection{Feature Ranking and Classification}
The discriminating characteristic of each feature is evaluated using 3 ranking methods. First, the F-score of each feature ($\phi$) is evaluated using (\ref{feqn}). Here, for $c$ different class labels, the mean feature value ($v$) for all samples in class $c$ is denoted as $\overline{v^{c}_{\phi}}$, while the overall mean feature value is $\overline{v_{\phi}}$. The number of samples belonging to each class type is $n^{c}$ and total number of samples is $n$. The second feature ranking method utilizes the correlation coefficient between feature distributions as a metric for feature ranking in (2). Here, the underlying assumption is that discriminating characteristic of a feature ($\phi_1$) can be improved by using it in combination with other strongly correlating features ($\phi_2$). Thus, features are ranked in the decreasing order of their correlation coefficients ($\rho$) with the remaining features using (2). The third feature ranking strategy uses mRMR criterion \cite{Major} that is based on mutual information from the individual features. Here, features are ranked based on the top combination of features that have maximum relevance with the sample class labels and minimum redundancy.

\begin{eqnarray}\label{feqn}
\forall \phi, F(_\phi)=\frac{\sum_{j=0}^{c-1} (\overline{v^{j}_{\phi}}-\overline{v_{\phi}})^2}{\sum_{j=0}^{c-1} \frac{1}{n^c-1} \sum_{k=1}^{n^c}(v_{k,\phi}^{c}-\overline{v^{c}_{\phi}})^2}.\\
\rho(_{\phi_1},_{\phi_2})_=\frac{\overline{v_{\phi_1}v_{\phi_2}}-\overline{v_{\phi_1}}.\overline{v_{\phi_2}}}{\frac{1}{n}\sqrt{{\sum_{_{k=1}}^{n}(v_{k,\phi_1}-\overline{v_{\phi_1}})^2{\sum_{_{k'=1}}^{n}(v_{k',\phi_2}-\overline{v_{\phi_2}})^2}}}}.\\ \nonumber
\end{eqnarray}

\begin{figure}[ht]
\begin{center}
\includegraphics[width = 2.7in, height=2.0in]{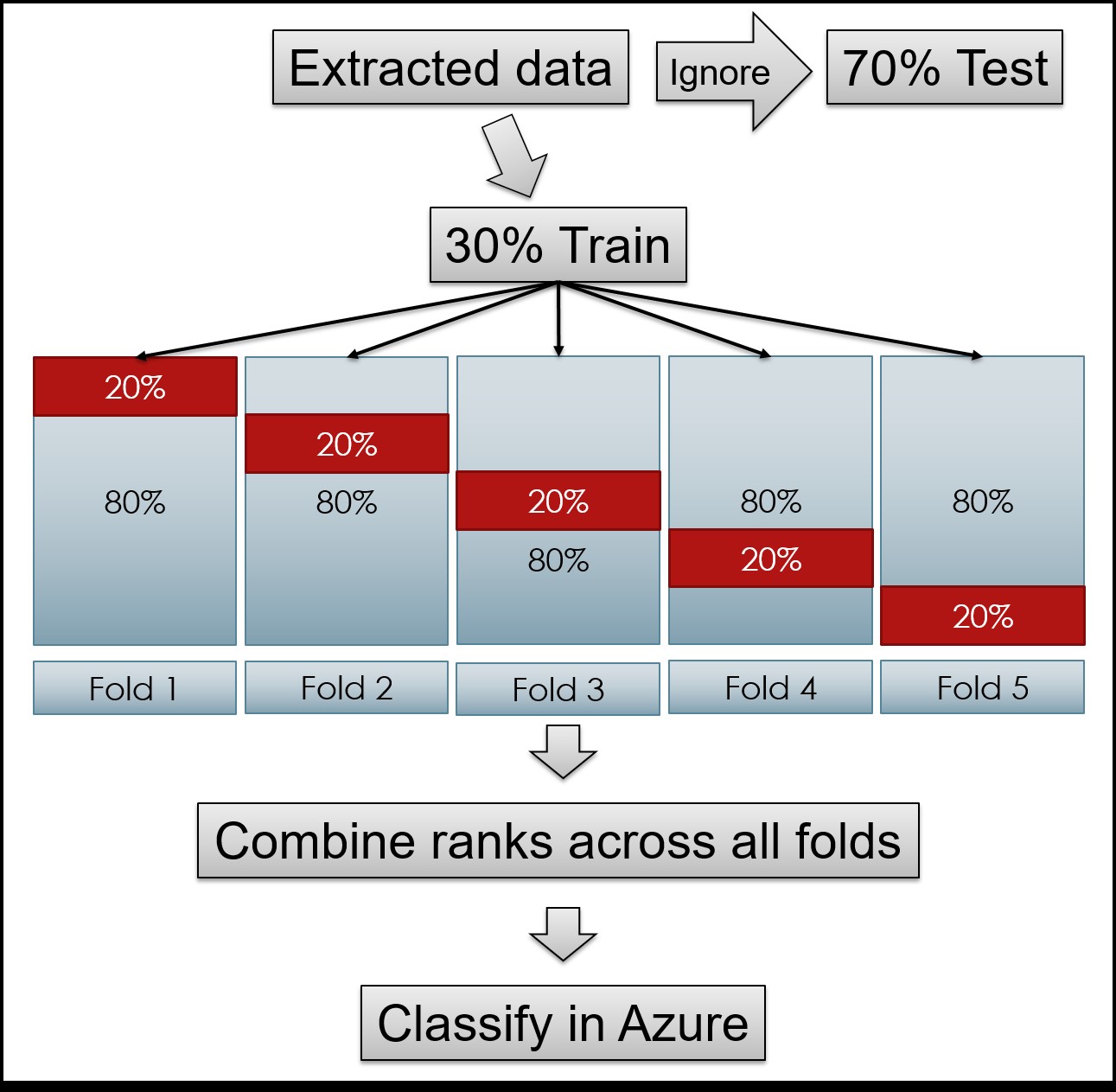}
\caption{Feature ranking process with 5-fold cross-validation.
}\label{fold}
\end{center}
\end{figure}

For optimal feature ranking, 5-fold cross validation followed by classification is performed. First, each data set is partitioned into training data (30\% samples) and testing data (70\% samples) \cite{DREAM}. Next, the training data set is separated into 5-folds, where in each fold, 80\% of the data samples are used for feature ranking and classifier parametrization, while the remaining 20\% data samples are used for validation of the trained classifier. The averaged ranks across all the folds are analyzed for aggregated classification performance as shown in Fig. \ref{fold}. 
Finally, optimal classifier selection is performed for the two data sets from a family of classifiers including k-nearest neighbors, Gaussian Mixture Models, Support Vector Machines, Decision Forest (DF), Boosted Decision Trees (BDT). It is observed that the BDT and DF classifiers have least average validation error for the DIARETDB1 and STARE data sets, respectively. 
\section{Results}\label{result}
\begin{figure}[ht]
\begin{center}
\subfigure[]{\includegraphics[width = 3.0in,height=1.55in]{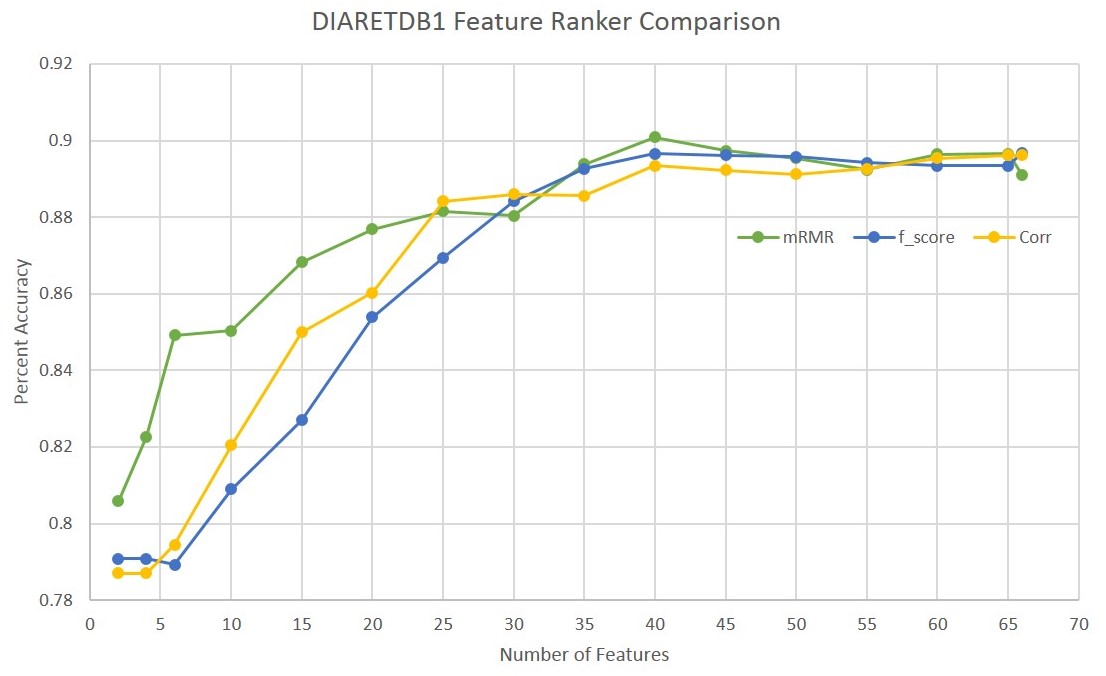}}
\subfigure[]{\includegraphics[width = 3.0in, height=1.55in]{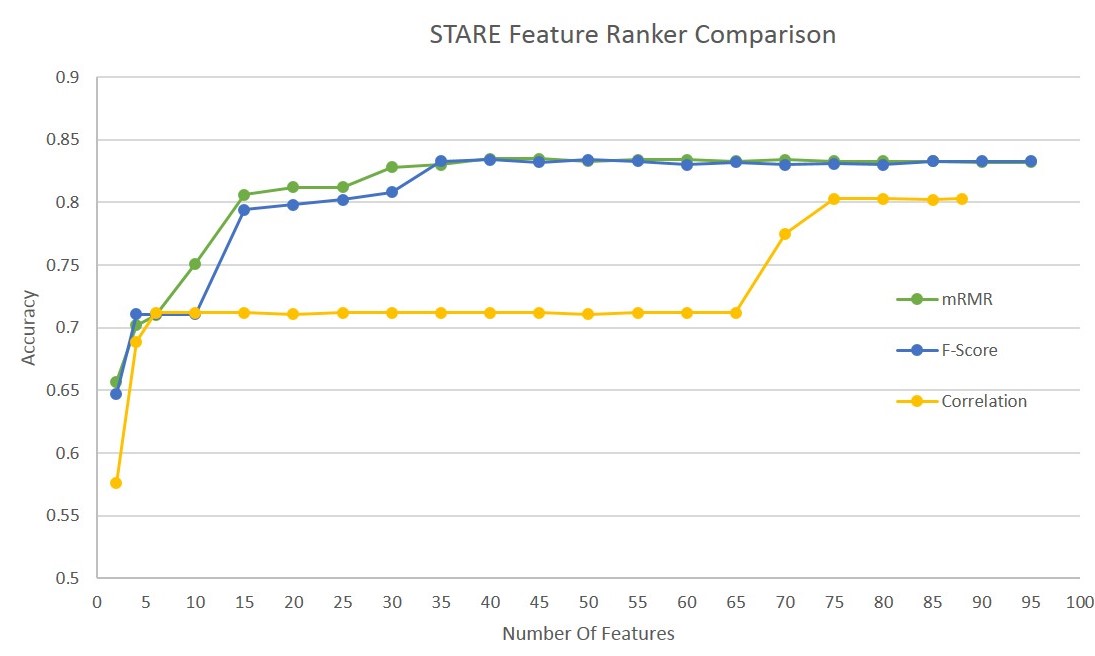}}
\caption{Classification performance assessment for top ranked features. Top 40 mRMR ranked features are most accurate for (a) DIARETDB1 data set, (b) STARE data set.}     
\label{feat}
\end{center}
\end{figure}

\begin{figure}
\begin{center}
\subfigure[]{\includegraphics[width = 1.5in, height=1.5in]{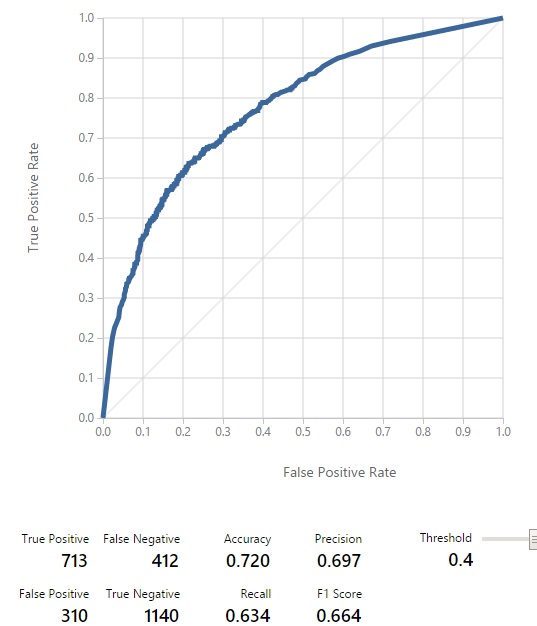}}
\subfigure[]{\includegraphics[width = 1.5in, height=1.5in]{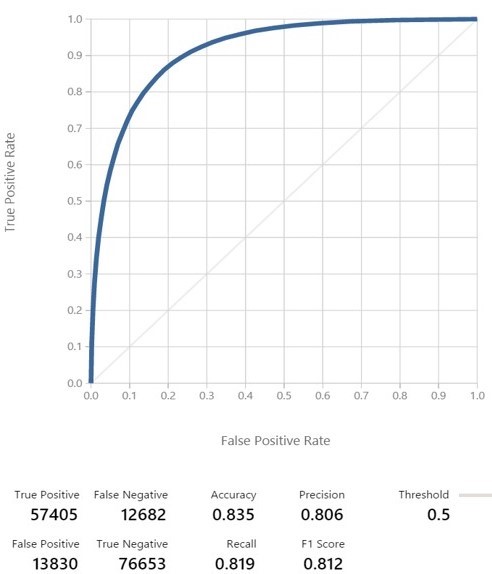}}
\caption{ROC curves generated by the MAMLS platform. (c) Classification of hemorrhages from microaneurysms in DIARETDB1 (AUC=0.78). (b) Classification of minor blood vessels in STARE (AUC=0.914).}     
\label{roc}
\end{center}
\end{figure}

In Fig. \ref{feat}, the average classification accuracy on the DIARETDB1 and STARE data sets are analyzed using the top $\phi$ combinations of ranked features, where $\phi \in [1:L]$, where $L=66$ and $L=98$ for the DIARETDB1 and STARE data sets, respectively. Here, it is observed that using the mRMR feature ranking strategy, the top 10-15 features are capable of achieving about 75-80\% classification accuracy, while the remaining 25-30 features contribute to an additional 3-6\% increase in overall classification accuracy. Thus, the top 10-15 features may be adequate for initial screening purposes, but the complete set of 40 features becomes important in case of borderline decision making tasks, i.e. separating fundus images with moderate NPDR from severe NPDR. 
For both DIARETDB1 and STARE data sets, the mRMR feature ranking strategy results in highest classification accuracy using top 40 features. For the DIARETDB1 data set, the top 40 features include the 14 structural, 11 Gaussian Coefficient, 9 regional intensity, 6 gradient intensity features from Table \ref{notation}. On this data set, the optimal feature set results in 1.2\% higher accuracy and 11.2\% lower computation time than the entire feature set. For the STARE data set, top 40 features include 4 pixel-window based, 4 pixel intensity-based, 14 structural, 10 Gaussian Coefficient, 8 regional intensity features from Table \ref{notation}. On this data set, the optimal feature set results in 0.24\% higher accuracy with 23.4\% lower processing time when compared to the entire feature set. The performance of the optimal feature set with respect to the existing methods is shown in Table \ref{res}.
 \begin{table*}[ht]
\begin{center}
\caption{Classification Accuracy of Optimal Feature Set in comparison with full feature set and existing works. Computation time is measured in the MAMLS platform.}
\begin{tabular}{|c| c| c| c| c|}
\hline 
Dataset&All Features(ACC)&Optimal Features (ACC)&Existing work/ Features (ACC)&Computation Time (seconds)\\ \hline\hline
DIARETDB1 \cite{DB1}&66 (0.89)&40 (0.901)&\cite{DREAM}/ 30 (0.886)&792 s\\ \hline
STARE \cite{STARE}&98 (0.832)&40 (0.835)& \cite{Major}/ 8 (0.751)&326 s\\ \hline
\end{tabular}
	\label{res}
	\end{center}
\end{table*}
The Receiver Operating Characteristic (ROC) curves and area under ROC curves (AUC) for the challenging classification tasks of hemorrhages from microaneurysms in the DIARTEDB1 data set and for classification of minor blood vessels from non-vessels is shown in Fig. \ref{roc}. Using the optimal set of top 40 features, the observed [sensitivity (SEN), specificity (SPEC), AUC] for classification of red lesions from false positive regions is [0.9,0.7,0.895], which has better DR screening performance than [0.8,0.85,0.84] reported in \cite{DREAM}.

\section{Conclusions and Discussion} \label{conclusion}
In this paper optimal feature sets have been identified for classification of NPDR lesions and minor vessels that can aid automated DR screening systems \cite{DREAM}. It is observed that mRMR feature ranking strategy is most efficient in detecting combination of region-based and pixel-based features for DR classification tasks. Additionally, Decision Forest and Boosted Decision Tree classifiers in the MAMLS platform were found to be most effective for such large-scale fundus image data classification. The data sets used for the proposed analysis are available for download and classification performance analysis \footnote[1]{https://sites.google.com/a/uw.edu/src/useful-links}. Future efforts will be directed towards evaluating the proposed large-scale screening systems for NPDR and PDR on additional fundus image data sets. 

\bibliographystyle{IEEEtran}
\bibliography{IEEEabrv,references}

\end{document}